\documentclass[11pt,letterpaper]{article}
\usepackage{emnlp2016}
\usepackage{times}
\usepackage{latexsym}
\usepackage{url}

\emnlpfinalcopy

\def\emnlppaperid{***}

\title{Large-Scale Machine Translation between Arabic and Hebrew:\\Available Corpora and Initial Results}

\author{Yonatan Belinkov \and James Glass \\
 MIT
  Computer Science and Artificial Intelligence Laboratory \\
  Cambridge, MA 02139, USA \\
  {\tt \{belinkov, glass\}@mit.edu}  }

\date{}

\begin{document}

\maketitle

\begin{abstract}
\vspace*{-5pt}
  Machine translation between Arabic and Hebrew has so far been limited by a lack of parallel corpora, despite the political and cultural importance of this language pair. Previous work relied on manually-crafted grammars or pivoting via English, both of which are unsatisfactory for building a scalable and accurate MT system. In this work, we compare standard phrase-based and neural systems on Arabic-Hebrew translation. We experiment with tokenization by external tools and sub-word modeling by character-level neural models, and show that both methods lead to improved translation performance, with a small advantage to the neural models.  
\end{abstract}

\vspace*{-10pt}

\section{Introduction}
Arabic and Hebrew are Semitic languages spoken by peoples with complicated cultural and political relationships. They share important similar characteristics in all linguistic levels, including orthography, morphology, syntax, and lexicon. Yet there is relatively little previous research on machine translation between the two languages, despite its potential benefit for promoting understanding between their speakers. The main reason for this lacuna is a lack of parallel Arabic-Hebrew texts. This has led researchers to consider alternative approaches, such as pivoting via English~\cite{elkholy2014alignment,el2015morphological} or developing transfer-based systems built with synchronous context free grammars~\cite{Shilon:2012:MTH:2159073.2159155}. Both approaches are unsatisfactory: the transfer-based system relies on manually-crafted grammars and lexicons, therefore suffering from robustness issues, and pivoting via a morphologically-poor language like English leads to under-specification of potentially useful features.

Recently, a number of large-scale parallel Arabic-Hebrew corpora have been compiled, mostly from multilingual transcriptions of spoken language available online~\cite{cettoloEtAl:EAMT2012,LISON16.947}. These resources finally allow for training full-scale statistical machine translation systems on the Arabic-Hebrew pair. Our first contribution is in evaluating such standard systems on a clearly-defined dataset. We compare phrase-based machine translation (PBMT) with neural machine translation (NMT), using state-of-the-art implementations. 

Like other Semitic languages, Arabic and Hebrew feature rich morphology and frequent cliticization (joining of prepositions, conjunctions, etc. to the main word). These characteristics lead to increased ambiguity and pose a challenge to machine translation. A common solution is to apply tokenization by external tools, shown to help translation between Arabic/Hebrew and English~\cite{ElKholy2012,singh2012}. Our second contribution is thus in evaluating tokenization by external tools for the Arabic-Hebrew language pair. We also experiment with character-level neural models that have recently become popular for dealing with morphologically-rich languages~\cite{KimAAAI1612489}.

In this work, we focus on Arabic-to-Hebrew translation. Arabic has relatively more available resources such as tokenizers and morphological analyzers, making this translation direction more approachable. We leave the investigation of Hebrew-to-Arabic translation for future work. 

Our results show that phrase-based and neural MT systems reach comparable performance, with a small advantage to neural models. We also ascertain the importance of sub-word modeling, where neural character models rival or surpass morphology-aware tokenization by standard tools. We conclude by pointing to potential directions for future research. 


\section{Related Work} \label{sec:related-work}
There is relatively little previous research on machine translation between Arabic and Hebrew, despite cultural and political relations between their speakers,
and despite their linguistic similarities. The most relevant work is by~\newcite{Shilon:2012:MTH:2159073.2159155}, who built a statistical transfer-based system for translating from Arabic to Hebrew and vice versa. Their work relies on synchronous context free grammars and lexicons in the two languages, an approach that they advocate as being better suited to this pair for two main reasons: (a) a lack of available parallel corpora; and (b) the rich morphology of Arabic and Hebrew that requires linguistic knowledge. Here, we explore an alternative to this approach by exploiting Arabic-Hebrew parallel texts that have recently become available, enabling us to train standard statistical MT systems.\footnote{\newcite{Cettolo:16} describes the corpus and baseline MT systems in work concurrent with this paper.} We further explore methods for handling morphology both by using traditional tools for morphological analysis and tokenization, and by training a \emph{character}-level neural MT system. 

Other work directly targeting machine translation between Arabic and Hebrew includes \cite{elkholy2014alignment}, which used pivoting via English. They improved translation quality by carefully designing the alignment symmetrization process in a phrase-based system. In later work, ~\newcite{el2015morphological} incorporated morphological constraints for pivoting in a phrase-based system, which they augmented with parallel Arabic-Hebrew data (from an earlier version of the corpus we use in this paper).  While pivoting is an appealing solution to scarcity in parallel corpora,  \newcite{Shilon:2012:MTH:2159073.2159155} convincingly show how pivoting through a morphologically-poor language like English leads to under-specification of linguistic features and loss of information. 

There is a fairly decent body of work on translation between Arabic and English, using a variety of methods; see the survey in~\cite{Alqudsi2014}. In particular, the importance of morphology-aware tokenization when translating from and to Arabic has been confirmed in phrase-based~\cite{Badr:2008:SES:1557690.1557732,Habash:2006:APS:1614049.1614062,ElKholy2012} and neural machine translation, in both hybrid~\cite{devlin-EtAl:2014:P14-1} and end-to-end systems~\cite{almahairi2016first}. 
Work on Hebrew translation is more limited, but previous studies on translating Hebrew to English also demonstrated the need for morphological analysis and tokenization~\cite{lavie2004rapid,lembersky-J12-4004,singh2012}. 



\section{Linguistic Description}
We give here a short description of similarities and differences between Arabic and Hebrew, referring to \cite{Shilon:2012:MTH:2159073.2159155} for a  comprehensive discussion. 

As Semitic languages, Arabic and Hebrew share several characteristics. Both orthographies commonly omit vowels and other diacritics in writing, leading to increased ambiguity. The scripts are distinct, but there is substantial overlap in the alphabets. Many clitics (prepositions, conjunctions, definite articles) are prefixed or suffixed to words. Both languages have a rich morphology with a complex system of verbal inflection. Their inflection paradigms partially, but not completely, overlap. Syntactically, the languages have both verbal and verbless sentences. Arabic, in particular, has a more complicated agreement system. Some systematic word order patterns can be noted (SVO for Hebrew, VSO for Arabic), but these have exceptions and depend on genre.  

\newcite{Shilon:2012:MTH:2159073.2159155} discuss the challenges such characteristics pose for machine translation between Arabic and Hebrew. In this work, we mostly address orthographic and morphological challenges, which call for solutions like tokenization and representing sub-word elements. 

\section{Parallel Corpora} \label{sec:corpora}
Until recently, there were not many available parallel corpora of Arabic and Hebrew. \newcite{Shilon:2012:MTH:2159073.2159155} prepared a parallel corpus of several hundred sentences from the news domain, too small for training a statistical system but potentially useful for evaluation. Since then, two large resources have become available. First, WIT$^3$ provides multilingual transcriptions of TED talks~\cite{cettoloEtAl:EAMT2012} and its 2016 release includes about 3 million words of Arabic-Hebrew parallel texts~\cite{Cettolo:16}. 
As a corpus of TED talks, it has several interesting features: diversity of topics, spoken language transcriptions, and user-generated translations, although the review process ensures a reasonable translation quality. The original transcriptions are segmented at the caption level and WIT$^3$ automatically joins them into sentences.

Second, OPUS provides a collection of translation texts  from the web. 
The largest Arabic-Hebrew parallel corpus is  OpenSubtitles, comprising automatically aligned movie and TV subtitles. The 2016 release contains more than 100 million words~\cite{LISON16.947}. In addition, OPUS provides a version with alternative translations, with some 70 million words of Arabic-Hebrew texts~\cite{TIEDEMANN16.62}. Having alternative translations can be valuable for evaluation with multiple references, although many alternatives are simply duplicates. While this is by far the largest available Arabic-Hebrew parallel corpus, it suffers from the usual problems of OpenSubtitles texts: user-generated content, questionable translation quality, and automatic caption alignment. In addition, the right-to-left scripts cause problems with punctuation marks such as misplacement and wrong tokenization.

\begin{table}[t]
\footnotesize
\centering
\begin{tabular}{|l|l|l|l|}
\multicolumn{1}{c}{Corpus} & \multicolumn{1}{c}{Sents} & \multicolumn{1}{c}{Ar words} & \multicolumn{1}{c}{He words}  \\
\hline
OpenSubtitles & 14.6M & 108M & 111M  \\
OpenSubtitles-Alt & 9.5M & 71M & 76M  \\
WIT$^3$ & 0.2M & 3.4M & 3.1M  \\
GNOME & 0.6M & 2.1M & 2.6M  \\
KDE & 80.5K & 0.5M & 0.4M \\
Ubuntu & 51.3K & 0.2M & 0.2M  \\
Shilon et al. & 1.6K & 28K & 25K  \\
Tatoeba & 0.9K & 90K & 0.6M  \\
GlobalVoices & 76 & 3.2K & 3.7K  \\
\hline
\end{tabular}
\caption{Statistics of parallel Arabic-Hebrew corpora. See text for references and more details.}
\label{tab:corpora}
\vspace{-10pt}
\end{table}

Smaller Arabic-Hebrew corpora in OPUS include  localization files (Ubuntu, KDE, GNOME), 
each totaling between 200 thousand to 2 million words, as well as 
user-contributed translations from Tatoeba, and news stories from  GlobalVoices~\cite{Tiedemann:RANLP5,TIEDEMANN12.463}. 
\mbox{Table~\ref{tab:corpora}} summarizes statistics about available Arabic-Hebrew corpora.

\section{Experimental Setup}
\subsection{Machine Translation Systems} 
\paragraph{Phrase-Based MT}
We build a standard PBMT system using Moses~\cite{Koehn:2007:MOS:1557769.1557821}. Word alignment is extracted by \texttt{fast\_align}~\cite{dyer-chahuneau-smith:2013:NAACL-HLT} and symmetrized with the grow-diag-final-and strategy, and lexical reordering follows the msd-bidirectional-fe configuration. Sentences longer than 80 words are filtered during training. We train a 5-gram language model on the training set target side using KenLM~\cite{Heafield-estimate} and tune with MERT to optimize BLEU. These are common Moses settings that have also been used in Arabic-English translation~\cite{almahairi2016first}. 

\paragraph{Neural MT}
We train a neural translation system using a Torch~\cite{collobert2011torch7} implementation of attention sequence-to-sequence learning~\cite{kim2016}. We keep the default settings and experiment with two architectures: a small 2-layer 500 unit LSTM (on both encoder and decoder sides) and a larger 4-layer 1000 unit LSTM. 
Sentences are limited to 50 words and the vocabulary size is limited to 50,000 on both source and target sides. 
The model is trained on a single GPU using SGD. Decoding is done with beam search and a width of 5.  

\subsection{Tokenization and Sub-Word Models}
Morphological processing and tokenization are considered crucial for machine translation from and to Semitic languages like Arabic and Hebrew (Section~\ref{sec:related-work}). This is typically applied as a preprocessing step, requiring language-specific tools. An alternative option is to incorporate language-agnostic sub-word elements inside the training algorithm. We describe the two options next.  

\paragraph{Tokenization}
We experiment with tokenization of the Arabic source side using two tools: MADAMIRA~\cite{PASHA14.593}, a standard morphological analyzer and disambiguator, and the Farasa segmenter~\cite{abdelali-EtAl:2016:N16-3}, a much faster ranker that has been shown to perform comparably to MADAMIRA. In both cases we segment the Arabic according to the ATB scheme that tends to perform better than other schemes in translating between Arabic and English~\cite{ElKholy2012,sajjad:2013:qcri}. This scheme separates all clitics other than the definite article. While it is possible that other schemes will work better for Arabic-Hebrew translation, exploring this option is left for future work. The tokenized text is also normalized with the tools' default settings. On the Hebrew side, we only separate punctuation marks. 

\paragraph{Character-level models}
Character-level models have been shown to benefit neural MT, especially for languages with large vocabularies. For instance, \newcite{SennrichP16-1162} convert words to sub-word elements using byte-pair encoding and obtain significant gains on English-German/Russian translation. The method was also applied to Arabic-English translation~\cite{abdelali-EtAl:2016:N16-3}. Here we experiment with a character-level convolutional neural network (charCNN) that replaces input word vectors with learned representations based on character vectors~\cite{KimAAAI1612489}. We use the default settings in \cite{kim2016}. 

\subsection{Data and Evaluation}
We mainly experiment with the WIT$^3$ corpus of TED talks (Section~\ref{sec:corpora}). It is a fairly large corpus (3 million words), with high-quality translations and diverse topics. We use the designated train.tags files for training, IWSLT16.TED.tst2010-2014 for tuning, and IWSLT16.TED.dev2010 for testing. We keep IWSLT16.TED.tst2015-2016 as a held-out set for future evaluations. 
Table~\ref{tab:stats-wit3} provides some statistics about the datasets. 

\begin{table}[t]
\footnotesize
\centering
\begin{tabular}{|l|lll|}
\multicolumn{1}{l}{} & \multicolumn{1}{l}{Train} & \multicolumn{1}{l}{Tune} & \multicolumn{1}{l}{Test} \\
\hline
Sents & 0.2M & 7.3K & 874 \\
Ar words & 3.2M & 102.2K & 13.7K \\
He words & 3.0M & 93.1K & 12.7K \\
\hline
\end{tabular}
\caption{Number of sentences and (space-delimited) words in the WIT$^3$ corpus of TED talks used in our experiments.}
\label{tab:stats-wit3}
\end{table}

We also performed initial separate experiments with the OpenSubtitles corpus. 
However, the translation quality was very poor, mostly due to the extremely noisy nature of the dataset. Therefore we leave the exploration of this corpus for future work. 

We compute BLEU scores using the \texttt{multi-bleu.perl} script included with Moses. Significance testing follows \cite{koehn:2004:EMNLP,riezler2005some}. We also report Meteor scores (version 1.5), using Meteor Universal~\cite{denkowski:lavie:meteor-wmt:2014} to build language resources based on the phrase table learned by the PBMT system.  


\section{Results}
Table~\ref{tab:results-wit3} summarizes the results for Arabic-to-Hebrew translation on the WIT$^3$ corpus of TED talks. As expected, tokenization helps phrase-based MT, although the differences in BLEU scores are not statistically significant. In terms of BLEU, neural MT performs significantly better than phrase-based MT, and char-based models lead to substantial and statistically significant improvement. Another small improvement is gained by replacing generated unknown words with translations of their aligned source words based on the attention weights~\cite{jean-EtAl:2015:ACL-IJCNLP}. Using a larger and deeper NMT model does not lead to significant improvement, possibly due to the size of the training data.

\begin{table}[t]
\footnotesize
\centering
\begin{tabular}{|l|l|l|l|}
\multicolumn{1}{l}{System} & \multicolumn{1}{c}{BLEU} & \multicolumn{1}{c}{Meteor} & \multicolumn{1}{c}{PPL} \\
\hline
PBMT & 9.31 & 32.30 & 478.4 \\
PBMT+Tok-Farasa & 9.51 & 33.38 & 335.5 \\
PBMT+Tok-MADAMIRA & 9.63 & 32.90 & 342.5 \\
\hline
NMT & 9.91 & 30.55 & 2.275 \\
NMT Large & 9.92 & 30.46 & 2.214 \\
NMT+UNK Replace & 10.12 & 31.84 & 2.275 \\
NMT+charCNN & 10.65 & 32.43 & 2.239 \\
NMT+charCNN+UNK Repl. & 10.86 & 33.61 & 2.239 \\
\hline
\end{tabular}
\caption{Results on WIT$^3$. Differences in BLEU scores in the first block are not statistically significant (at $p < 0.05$); differences between the two blocks are significant; the difference between small and large NMT models is not significant; differences between word and character NMT models are significant. Perplexity (PPL) scores are computed from the PBMT language model and the NMT decoder's classification loss, respectively.}
\label{tab:results-wit3}
\end{table}

We note that the generally low BLEU scores can be attributed to the single-reference evaluation mode, as well as the challenging nature of the data (spoken language transcripts, automatically aligned captions, diverse topics). Similar BLEU scores were reported for translating from English into Arabic and Hebrew in previous evaluations of TED talks translation~\cite{cettolo2014report}. 

Looking at Meteor scores, we again see that tokenization helps, but this time the basic NMT system is inferior to PBMT. However, as Meteor Universal uses the phrase-table learned by the PBMT system, it might be biased towards PBMT. Using a character-based model and UNK replacement can close this gap, leading to the best performing system.



\section{Conclusion and Future Work}
We presented initial experiments in large-scale Arabic-to-Hebrew machine translation, comparing both phrase-based and neural MT. We also evaluated the contribution of tokenization to the PBMT system and of character-level models to the NMT system. 

This work is a first step that can be extended in a number of ways. First, experimenting with the Hebrew-to-Arabic direction might reveal new insights. Second, other combinations of tokenization and character-level models can be explored (e.g. character-level neural models on tokenized or byte-pair encoded text). The parallel corpora can also be cleaned and improved, especially by adding multiple reference translations. Finally, modeling inter-relations between the two languages in a more direct manner is an appealing direction, given the similarities across linguistic levels. 

\section*{Acknowledgments}
The authors would like to thank Mauro Cettolo for useful suggestions with regards to the Arabic-Hebrew TED talks corpus, Pierre Lison and J\"org Tiedemann for help with getting access to the OpenSubtitles corpus, and Reshef Shilon for fruitful discussions. 
This work was supported by the Qatar Computing Research Institute (QCRI). Any opinions, findings, conclusions, or recommendations expressed in this paper are those of the authors, and do not necessarily reflect the views of the funding organizations.

\bibliography{semat2016}
\bibliographystyle{emnlp2016}

\end{document}